\documentclass[10pt,twocolumn,letterpaper]{article}

\usepackage[pagenumbers]{cvpr} 

\usepackage{graphicx}
\usepackage{amsmath}
\usepackage{amssymb}
\usepackage{booktabs}
\usepackage{makecell}
\usepackage{multirow}
\usepackage[accsupp]{axessibility}

\usepackage[pagebackref,breaklinks,colorlinks]{hyperref}

\usepackage[capitalize]{cleveref}
\crefname{section}{Sec.}{Secs.}
\Crefname{section}{Section}{Sections}
\Crefname{table}{Table}{Tables}
\crefname{table}{Tab.}{Tabs.}

\def\cvprPaperID{6671} 
\def\confName{CVPR}
\def\confYear{2023}

\begin{document}

\title{VLPD: Context-Aware Pedestrian Detection \\via Vision-Language Semantic Self-Supervision}

\author{
Mengyin Liu\textsuperscript{1*} \and Jie Jiang\textsuperscript{2*} \and  Chao Zhu\textsuperscript{1$\dagger$} \and Xu-Cheng Yin\textsuperscript{1} \and
\textsuperscript{1}School of Computer and Communication Engineering, \\
University of Science and Technology Beijing, Beijing, China\\
\textsuperscript{2}Data Platform Department, Tencent, Shenzhen, China \\
{\tt\small blean@live.cn, zeus@tencent.com, \{chaozhu, xuchengyin\}@ustb.edu.cn }
}

\maketitle

\let\thefootnote\relax\footnotetext{$*$ Equal contribution. $\dagger$ Corresponding author.}

\begin{abstract}
   Detecting pedestrians accurately in urban scenes is significant for realistic applications like autonomous driving or video surveillance. However, confusing human-like objects often lead to wrong detections, and small scale or heavily occluded pedestrians are easily missed due to their unusual appearances. To address these challenges, only object regions are inadequate, thus how to fully utilize more explicit and semantic contexts becomes a key problem. Meanwhile, previous context-aware pedestrian detectors either only learn latent contexts with visual clues, or need laborious annotations to obtain explicit and semantic contexts. Therefore, we propose in this paper a novel approach via Vision-Language semantic self-supervision for context-aware Pedestrian Detection (VLPD) to model explicitly semantic contexts without any extra annotations. Firstly, we propose a self-supervised Vision-Language Semantic (VLS) segmentation method, which learns both fully-supervised pedestrian detection and contextual segmentation via self-generated explicit labels of semantic classes by vision-language models. Furthermore, a self-supervised Prototypical Semantic Contrastive (PSC) learning method is proposed to better discriminate pedestrians and other classes, based on more explicit and semantic contexts obtained from VLS. Extensive experiments on popular benchmarks show that our proposed VLPD achieves superior performances over the previous state-of-the-arts, particularly under challenging circumstances like small scale and heavy occlusion. Code is available at \href{https://github.com/lmy98129/VLPD}{https://github.com/lmy98129/VLPD}.
\end{abstract}

\section{Introduction}
\label{sec:intro}
\begin{figure}
	\centering
	\includegraphics[width=0.981\linewidth]{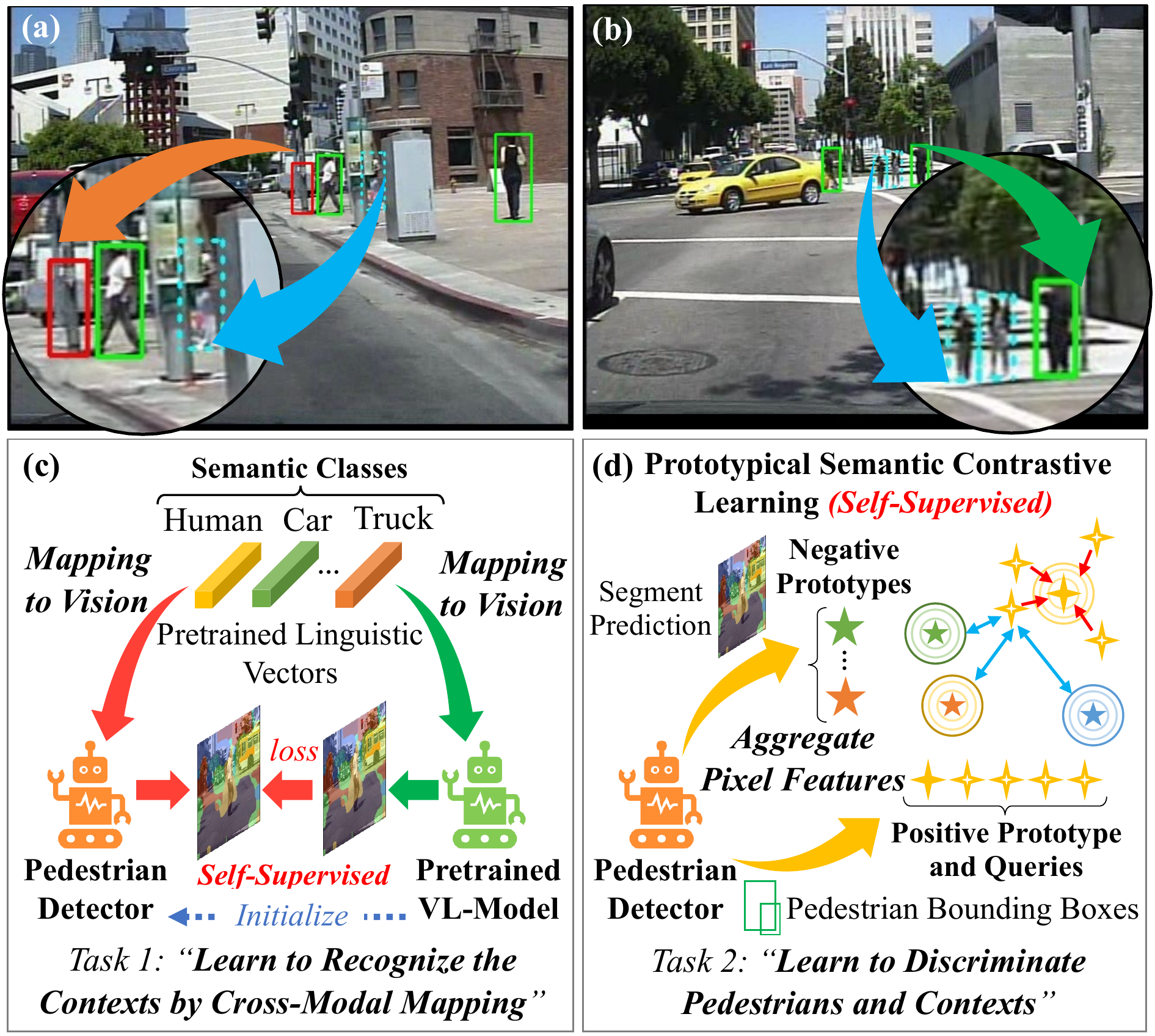}
	
	\caption{Illustration of the problems by previous works (top) and our proposed method to tackle them (bottom). (a) and (b) are predicted by \cite{liu2019high}. Green boxes are correct, red ones are human-like traffic signs, and dashed blue ones are missing heavily occluded or small scale pedestrians. (c) and (d): We propose self-supervisions to recognize the contexts and discriminate them from pedestrians.}
	\label{fig:Teaser}
\end{figure}

With the recent advances of pedestrian detection, enormous applications benefit from such a fundamental perception technique, including person re-identification, video surveillance and autonomous driving. In the meantime, various challenges from the urban contexts, i.e., pedestrians and non-human objects, still hinder the better performances of detection. For example, confusing appearances of human-like objects often mislead the detector, as shown in Figure \ref{fig:Teaser}(a). Moreover, heavily occluded or small scale pedestrians have unusual appearances and cause missing detections as Figure \ref{fig:Teaser}(a) and (b). Apart from the object regions, the contexts are crucial to address these challenges.

\begin{figure*}
	\centering
	\includegraphics[width=0.812\textwidth]{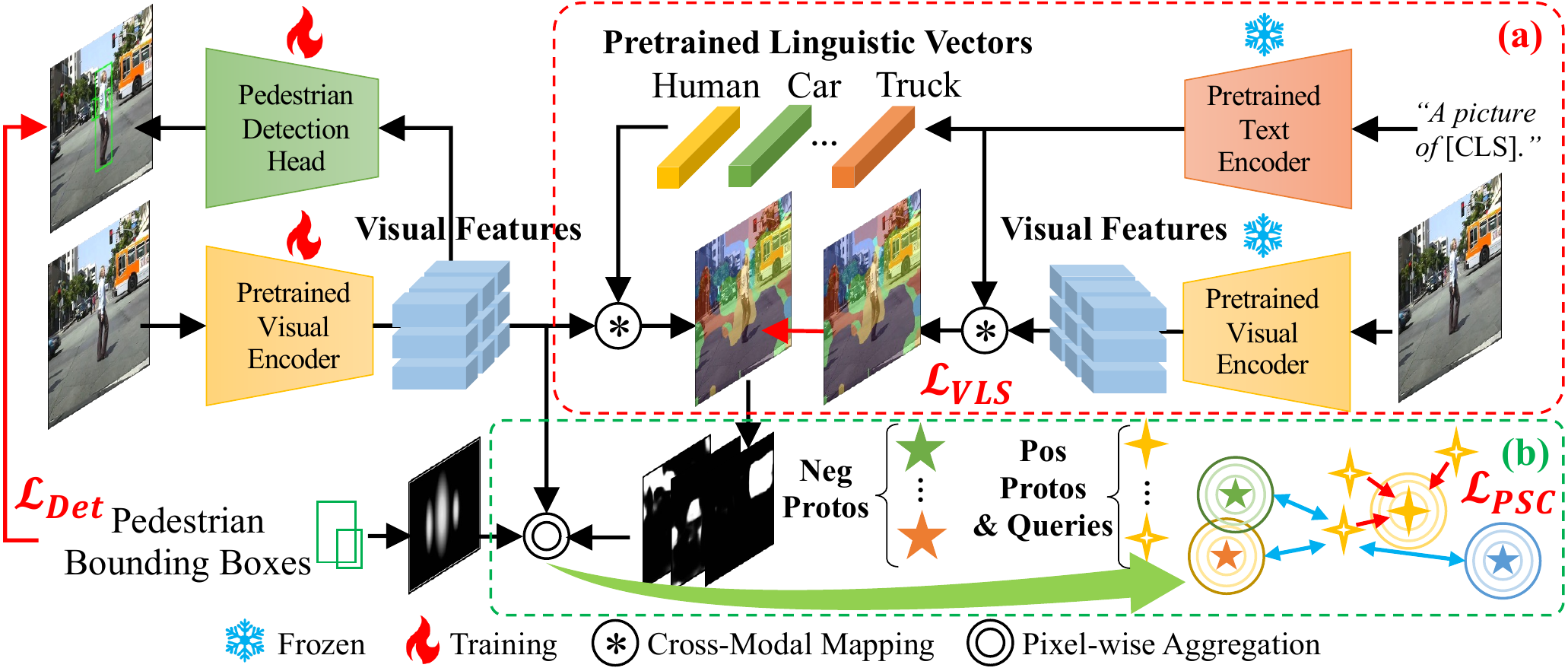}
	
	\caption{The overall architecture of our proposed VLPD approach. (a) Vision-Language Semantic (VLS) segmentation obtains pseudo labels via Cross-Modal Mapping, then the Pretrained Visual Encoder learns fully-supervised detection ($\mathcal{L}_{Det}$) and self-supervised segmentation to recognize semantic classes for explicit contexts without any annotations. (b) Prototypical Semantic Contrastive (PSC) learning lets the pixel-wise pedestrian features as queries closer to positive prototypes and further to negative ones based on Pixel-wise Aggregation.}
	\label{fig:Pipeline}
\end{figure*}

Nevertheless, previous methods still make inadequate investigations on the contexts in urban scenarios.  For instance, manual contextual annotations from CityScapes \cite{cordts2016cityscapes} boost SMPD \cite{jiang2022urban} on the  pedestrian benchmark CityPersons \cite{zhang2017citypersons}, because they share homologous image data.  Besides, a semi-supervised model yields pseudo labels for the Caltech dataset \cite{dollar2009pedestrian}. However, both these two solutions require expensive fine-grained annotations, especially for training the semi-supervised model. Moreover, other methods learn regional latent contexts merely from limited visual neighborhood \cite{zhang2020feature}, or non-human local proposals as negative samples for contrastive learning \cite{lin2022pedestrian}. Without an explicit awareness of semantic classes in the contexts, these methods thus still suffer from unsatisfactory performance. 

Besides, some pedestrian detection methods also indirectly handle the contexts. For the occlusion problems, many part-aware methods \cite{zhang2018occlusion,zhang2018occluded,chi2020pedhunter,xie2020mask,lu2020semantic,he2021occluded,song2022prnet++,li2022occluded,li2022oaf} adopt visible annotations for the occluded pedestrians, which indicate the occlusion by other pedestrians or non-human objects in the contexts. Whereas, these labels still need heavy labors of human annotators. For scale variation \cite{lin2018graininess,cao2019taking,zhang2020asymmetric,wu2020self,ding2021learning}, crowd occlusion \cite{liu2019adaptive,huang2020nms,zhou2020noh,xie2020count,wang2021mapd,zheng2022progressive} or generic hard pedestrians \cite{liu2018learning,liu2019high,brazil2019pedestrian,tesema2020hybrid,liu2021adaptive}, most previous works are intra-class, e.g., small pedestrians or crowded scenes, and thus irrelevant to context modeling problems.

Inspired by the vision-language models, we notice a more explicit context modeling without any annotations via cross-modal mapping. For instance, DenseCLIP \cite{rao2022denseclip} is initialized with vision-language pretrained CLIP model \cite{radford2021learning} to learn cross-modal mapping from pixel-wise features to linguistic vectors of human-annotated classes. Meanwhile, MaskCLIP \cite{zhou2022extract} generates pseudo labels via cross-modal mapping and train another visual model. Hence, complementing the initialized mapping and pseudo labeling, we propose to recognize the semantic classes for explicit contexts via self-supervised Vision-Language Semantic (VLS) segmentation, as shown in Figure \ref{fig:Teaser}(c) and \ref{fig:Pipeline}(a).

Furthermore, we consider that only pixel-wise scores are ambiguous to discriminate pedestrians and contexts. Due to the coarse-grained pseudo labels, some parts of pedestrians might have higher scores of other classes. Different from the regional contrastive learning \cite{lin2022pedestrian}, we introduce the concept of prototype \cite{wang2021exploring,zhou2022rethinking} for a global discrimination. Each pixel of pedestrian features is pulled closer to pixel-wise aggregated positive prototypes and pushed away from the negative ones of other classes based on the explicit contexts obtained from VLS.  As illustrated in Figure \ref{fig:Teaser}(d) and \ref{fig:Pipeline}(b), a novel contrastive self-supervision for pedestrian detection is proposed to better discriminate pedestrians and contexts. 

In conclusion, we have observed a dilemma between the heavy burden of manual annotation for explicit contexts and local implicit context modeling. Hence, we propose a novel approach to tackle these problems via \textbf{V}ision-\textbf{L}anguage semantic self-supervision for \textbf{P}edestrian \textbf{D}etection (\textbf{VLPD}). The main contributions of this paper are as follows: 

\begin{itemize}
	\item Firstly, the Vision-Language Semantic (VLS) segmentation method is proposed to model explicit semantic contexts by vision-language models. With pseudo labels via cross-modal mapping, the visual encoder learns fully-supervised detection and self-supervised segmentation to recognize the semantic classes for explicit contexts. \textbf{To our best knowledge, this is the first work to propose such a vision-language extra-annotation-free method for pedestrian detection}. 
	
	\item Secondly, we further propose the Prototypical Semantic Contrastive (PSC) learning method to better discriminate pedestrians and contexts.  The negative and positive prototypes are aggregated via the score maps of contextual semantic classes obtained from VLS and pedestrian bounding boxes, respectively. Each pixel of pedestrian features is pulled close to positive prototypes and pushed away from the negative ones, in order to strengthen the discrimination power of the detector.
	
	\item Finally, by the integration of VLS and PSC, our proposed approach VLPD achieves superior performances over the previous state-of-the-art methods on popular Caltech and CityPersons benchmarks, especially on the challenging small scale and occlusion subsets.
	
\end{itemize}

\section{Related Works}

\subsection{Pedestrian Detection}

In realistic applications, various circumstances are challenging for pedestrian detection, including occlusion, scale-variation and generic hard pedestrian handling. Here, we discuss these common problems as well as the context-aware methods which are specialized for these problems.

\subsubsection{Occlusion Handling}

As a research hot-spot of pedestrian detection, handling occlusion should make the best of limited information from visible parts of pedestrian, and also avoid the noisy one from occlusion by other pedestrians or non-human objects. 

On the one hand, part-aware methods handle the visible parts with other parts occluded by contextual objects. For example, OR-CNN \cite{zhang2018occlusion} re-scores parts to highlight the visible ones. PRNet++\cite{song2022prnet++} progressively refines the predicted visible and full-body boxes. Extra labels of less frequently occluded heads facilitate HBAN\cite{lu2020semantic}, JointDet \cite{chi2020relational} and PedHunter \cite{chi2020pedhunter}. Some methods \cite{cao2019taking,li2022oaf,li2022occluded} handle the visible and full bodies by parallel branches. Moreover, DMSFLN \cite{he2021occluded} explores the feature distributions between both branches.

On the other hand, crowd-aware methods are specialized to intra-class occlusion without context modeling. Some post-processing methods \cite{liu2019adaptive,huang2020nms,luo2021nms} focus on the over-suppression of dense predictions in crowd scenes, and the others \cite{zheng2022progressive} handle the under-suppression of sparse ones. For the heavily overlapped pedestrians, loss-based methods \cite{wang2021mapd,wang2018repulsion,xie2020count} identify them by learning representations.

Differently, our proposed VLPD discriminates non-human occluders and pedestrians via the contrastive learning of PSC, on the basis of the self-supervised learning via self-generated explicit labels of semantic classes from VLS.

\subsubsection{Scale-Variation Handling}

Scale-variation is another problem that potentially related to modeling the context. Due to the distance, the blurry and noisy appearances of both small pedestrians and non-human objects often confuse the detectors. Multiple branches \cite{lin2018graininess,zhang2020asymmetric,ding2021learning} are popular for modeling different scales. With the powerful FPN \cite{lin2017feature} architecture, LBST \cite{cao2019taking} detects smaller pedestrians with the fusion of bottom-up and top-down features. Differently, SML \cite{wu2020self} pushes the features of small-scaled pedestrians towards the distribution of large ones. 

Unfortunately, these works focus on the small-scale pedestrians, due to no labels for small non-human objects. Hence, our proposed VLPD uses label-free explicit contexts including the latter and performs better on different scales. 

\subsubsection{Generic Hard Pedestrian Handling}

The central issue to handle hard pedestrians is accurate localization. Plenty of previous works \cite{liu2018learning, brazil2019pedestrian} introduce multi-phase spatial refinements. Following the anchor-free style from generic object detection, the CSP \cite{liu2019high,li2020box} series decrease hyper-parameters with an adaptive prediction. AP$^2$M \cite{liu2021adaptive} matches proper parameters for different hard samples. 

For the context modeling, SMPD\cite{jiang2022urban} adopts extra segmentation annotations, EGCL\cite{lin2022pedestrian} uses contrastive learning by local proposals, and FC-Net\cite{zhang2020feature} learns latent features of the local contexts. Without any extra labels, our proposed VLPD can recognize explicit contextual objects via pseudo labels of VLS, and then discriminates them with pedestrians via more global positive and negative prototypes of PSC. 

\subsection{Segmentation by Vision-Language Pretraining}

Recent progress of vision-language pretraining CLIP \cite{radford2021learning} has facilitated more powerful segmentation methods. For example, cosine similarity, i.e., cross-modal mapping, is calculated between visual features and linguistic vectors to obtain segmentation results. DenseCLIP \cite{rao2022denseclip} and LSeg \cite{li2022language} initialize the model with a pretrained CLIP visual encoder, and then learn mapping features to annotated classes via linguistic vectors. For self-supervision, MaskCLIP \cite{xie2020mask} obtains the pseudo labels via the mapping and learns a new vision model, which is evaluated to be sub-optimal by \cite{rao2022denseclip}.

Differently, complementing the cross-modal mapping and pseudo labeling, our proposed novel self-supervised VLS recognizes semantic classes as explicit contexts without any extra labels for context-aware pedestrian detection.

\subsection{Prototypical Contrastive Learning}

Due to the spatial resolution of images, purely pixel-wise dense contrastive learning \cite{wang2021dense} leads to heavy computational burden, and only discriminates locally regardless of the global image. Hence, previous works introduce ``Prototypes'' \cite{wang2021exploring,zhou2022rethinking} as the alternatives for pixel features of each semantic class. Differently, our proposed PSC maintain the pixels of pedestrians as queries to keep their inner variance, which learns better discrimination between pedestrians and other classes for the contextual-aware pedestrian detection.

\section{Proposed Method}

\begin{figure}
	\centering
	\includegraphics[width=1.0\linewidth]{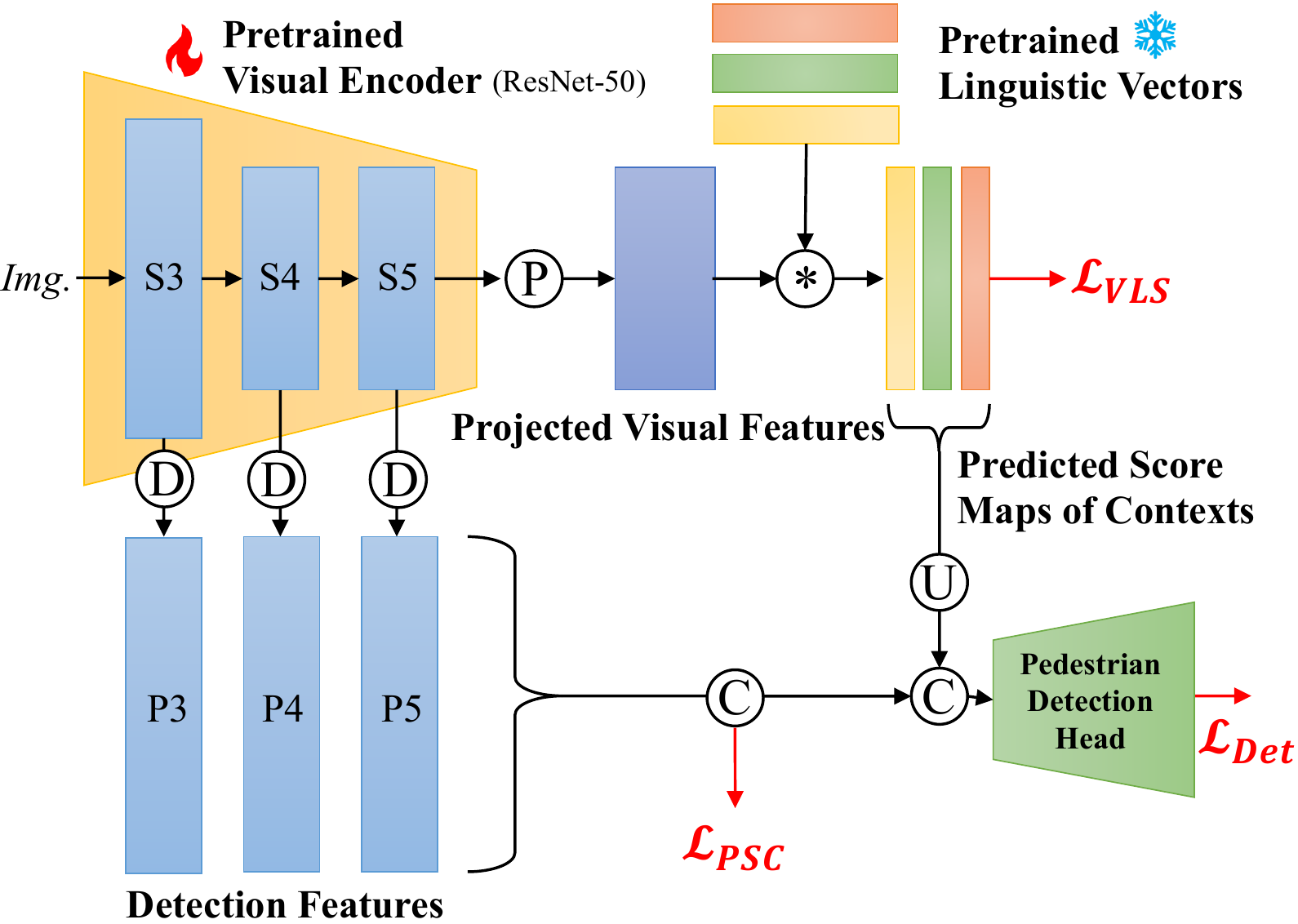}
	
	\caption{The detailed network architectures of our proposed VLPD. The visual encoder is ResNet-50 \cite{he2016deep} from the vision-language pretrained CLIP model \cite{radford2021learning}. Following the baseline CSP \cite{liu2019high}, S3$\sim$S5 are Deconvolved (``D'') as ``Detection Features'' for $\mathcal{L}_{Det}$ , which are further supervised by our $\mathcal{L}_{PSC}$. The Projections (``P'') before cross-modal mapping are adopted like \cite{zhou2022extract}. Concatenation and Up-sampling (``C'' and ``U'') are used for prediction as \cite{rao2022denseclip}. These details are omitted in Figure \ref{fig:Pipeline} for simplicity.}
	\label{fig:Details}
\end{figure}

As illustrated in Figure \ref{fig:Pipeline} and \ref{fig:Details}, our proposed approach \textbf{V}ision-\textbf{L}anguage semantic self-supervision for \textbf{P}edestrian \textbf{D}etection (\textbf{VLPD}) is an anchor-free detection framework following the baseline CSP \cite{liu2019high}. 
A pretrained visual encoder extracts features at different stages from S3 to S5. As shown in Figure \ref{fig:Details}, they are concatenated into ``Detection Features'' for the Detection Head to make predictions. 

To achieve the explicit semantic context modeling without any extra labels, our architecture comprises two key components: Vision-Language Semantic (VLS) segmentation and Prototypical Semantic Contrastive (PSC) learning. VLS leverages vision-language models to recognize the explicit contexts, where the visual encoder learns both fully-supervised pedestrian detection and segmentation via self-generated explicit labels of semantic classes by cross-modal mapping. PSC supervises the detector to better discriminate pedestrians and contextual semantic classes based on VLS. More details will be introduced in the following sections.

\subsection{Vision-Language Semantic Segmentation}

Benefiting from self-supervised cross-modal contrastive learning, vision-language models map the visual and linguistic vectors with similar meanings closer to each other into a unified feature space. Thus, it is possible to obtain the existences of semantic classes in an image via linguistic vectors. However, previous works initialize a model with cross-modal mapping for full-supervision \cite{rao2022denseclip,li2022language}, or re-train a new model by pseudo labels via the mapping \cite{xie2020mask}. 

Therefore, as shown in Figure \ref{fig:Pipeline}(a), we propose Vision-Language Semantic (VLS) segmentation as the complement of both initialized mapping and pseudo labeling.  Labels are generated by frozen pretrained models based on cross-modal mapping, thus the unfrozen visual encoder is supervised to predict the segmentation of explicit semantic classes, which serve as more global contexts rather than previous local latent ones \cite{zhang2020feature,lin2022pedestrian}. More details of our proposed VLS will be provided in the following sections.

\subsubsection{Cross-Modal Mapping for Pseudo Labeling}

\begin{figure}
	\centering
	\includegraphics[width=1.0\linewidth]{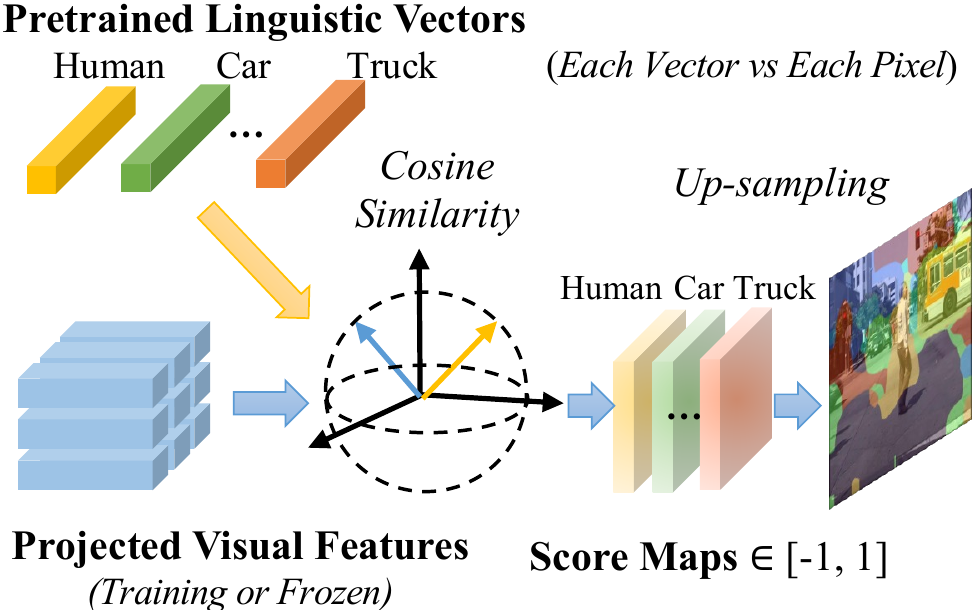}
	
	\caption{Cross-Modal Mapping for our proposed VLS. Visual features are projected as \cite{zhou2022extract} rather than a global vector in CLIP pretraining \cite{radford2021learning}. Cosine similarity is computed between each linguistic vector of semantic classes and each pixel of visual features, which constitutes the score maps of these classes in contexts.}
	\label{fig:Mapping}
\end{figure}

As one of the most popular vision-language models, CLIP \cite{radford2021learning} is capable of mapping image and text with similar meanings into closer vectors, based on its visual and linguistic encoders pretrained by self-supervised cross-modal contrastive learning. Although it is impossible to recover pixel-wise contextual information of an image from its visual vector after an attention-based pooling of CLIP \cite{radford2021learning}, this pooling operation can be modified into projections to keep visual regions, following \cite{zhou2022extract}. As shown in Figure \ref{fig:Mapping}, cosine similarities $S^c_{i}=((L^c)^{\top}(V_i))/(\Vert L^c \Vert \Vert V_i \Vert) \in S$ are calculated between pretrained linguistic vector $L^c\in\mathbb{R}^{\rm D'}$ of each class $c \in C$ and the projected vision features $V_i\in\mathbb{R}^{\rm D'}$ of each pixel $i=1,2,...,\mathrm{H' W'}$, which means the existence of each contextual class at each pixel of an image.

To obtain the pseudo labels for self-supervised learning, as illustrated in Figure \ref{fig:Pipeline}(a), images are feed into the frozen CLIP visual encoder to obtain visual features, and the linguistic vectors of classes are generated by frozen text encoder via the prompted sentence ``\textit{A picture of} [CLS]''. 

\subsubsection{Self-Supervised Learning for VLS}

Evaluated by the experiments of \cite{rao2022denseclip}, initialization with CLIP pretrained visual encoder contributes the maintenance of Cross-Modal Mapping, which allows the model to fully utilize the similarity between each pixel and each linguistic vector. While the ImageNet \cite{deng2009imagenet} pretrained one as \cite{zhou2022extract} needs re-training for an adaption to the mapping and thus is sub-optimal. Therefore, we propose to embrace both the advantage of both pseudo labeling and initialized mapping. 

In details, as shown in the Figure \ref{fig:Pipeline}(a) and \ref{fig:Details}, the visual encoder of pedestrian detector is initialized by CLIP pretrained parameters, which is identical to the frozen visual encoder for pseudo labeling as a self-supervision. Since the Cross-Modal Mapping is also performed, the linguistic vectors can be generated just once. The predicted $\bar{S}  \in \mathbb{R}^{\rm H'\times W' \times N}$ are supervised by pseudo labels $S$ based on Smooth L1 Loss \cite{girshick2015fast} $\mathcal{L}_{VLS}$ that is robust to noisy labels: 
\begin{equation}
	\mathcal{L}_{VLS} = \frac{1}{\mathrm{H' W' N}} \sum\nolimits_{i,c} Smooth\mathrm{L}1(\bar{S}^c_i,S^c_i),
	\label{eq:VLS}
\end{equation}
where class count $\mathrm{N}=|C|$. More than the self-supervision in Eq.\ref{eq:VLS}, the visual encoder is also trained to detect pedestrians as the baseline \cite{liu2019high}. Consequently, the model learns to model the contexts during detecting within a unified pipeline. The predicted $\bar{S}$ are fed into the Detection Head in Figure \ref{fig:Details} for an explicit contextual reference like \cite{rao2022denseclip}.

\subsubsection{Compacted Class Policy}

The complicated environments in the urban scenarios for pedestrian detection require a proper contextual class set. Meanwhile, urban dataset CityScapes \cite{cordts2016cityscapes} already defines various classes. However, experiments (in the following sections) reveal a negative effect of it. Inspired by imbalanced class frequency statistics in the CityScapes paper\cite{cordts2016cityscapes}, we propose Compacted Class Policy in Table \ref{table:Policy} to decide whether classes should be kept, compacted or discarded.

For the most classes, we adopt the $\mathrm{2}^{\rm nd}$ level of classes in CityScapes higher than the original ones, as illustrated in Table \ref{table:Policy}. Furthermore, the higher variance inside the ``vehicle'' class leads to performance loss in experimental trials. Thus, neither the $\mathrm{1}^{\rm st}$ nor $\mathrm{2}^{\rm nd}$ level of classes is applicable.

Therefore, we make statistics on frequencies in pixel-wise annotations of CityScapes for the images shared with CityPersons \cite{zhang2017citypersons}, which are computed via not only pixel-wise counting but also re-weighting by image-wise occurrence times. Less frequent tail classes are omitted by thresholding and the head ones are kept as the bottom of Table \ref{table:Policy}. 

In conclusion, our proposed VLS leverages the powerful vision-language model CLIP \cite{radford2021learning} to perform self-supervision based on Cross-Modal Mapping and Compacted Class Policy, which obtains pseudo labels of semantic classes for explicit contexts to learn recognizing them during detection for better discriminations. 

\begin{table}
	\centering
	\caption{Compacted Class Policy for our proposed VLS. }
	\begin{tabular}{c|c|c}
		\toprule [1.2pt]
		Original $\rightarrow$ & Compacted & Used  \\ 
		\midrule[0.8pt]
		\{road, sidewalk\} & ground & \checkmark \\
		\{building, wall, fence\} & building & \checkmark \\
		\{vegetation, terrain\} & tree & \checkmark \\
		\{person, rider\} & human & \checkmark \\
		\{pole, traffic light, traffic sign\} & traffic sign & \checkmark \\
		\midrule[0.8pt]
		\multirow{3.5}*{\makecell[c]{\{car, bicycle, bus, truck,\\motorcycle, train\}} }
		& vehicle & \texttimes \\
		\cmidrule{2-3}
		& \makecell[c]{\{car, bicycle,\\ bus, truck\}} & \checkmark \\
		\bottomrule[1.2pt]
	\end{tabular}
	\label{table:Policy}
\end{table}

\subsection{Prototypical Semantic Contrastive Learning}

Due to the coarse-grained characteristics of pseudo labels by our proposed VLS, some visible parts of the pedestrians might have higher scores of other classes from VLS, which are annotated by the bounding boxes for the detection tasks. Since there are no manual annotations available, explicit refinement to the pseudo labels is rather difficult. 

Inspired by self-supervised contrastive learning \cite{wang2021dense,wang2021exploring,zhou2022rethinking} for discriminative representations of positive and negative samples, we introduce this powerful technique and propose a novel Prototypical Semantic Contrastive (PSC) learning, which learns a better discrimination of pedestrians and other semantic classes without any extra labels. 

In order to decrease the heavy computations by dense pixel-wise methods \cite{wang2021dense}, the concept of ``Prototype'' is embraced. It means a representative feature which represents all the features belong to same semantic class. In this paper, prototype of pedestrian is positive, and others are negative.

Take the negative prototypes as examples. The predicted score maps $\bar{S} \in \mathbb{R}^{\rm H' \times W' \times (N-1)}$ are adopted as the indicator of the spatial existences of all the non-human $\mathrm{N}-1$ classes, except the ``Human'' which is overlapped with pedestrian. Here, we denote $C$ as non-human classes for simplicity, where $|C|=\mathrm{N}-1$. $\dot{S} \in \mathbb{R}^{\rm H\times W \times (N-1)}$ are up-sampled from $\bar{S}$. A SoftMax function $\delta$ with a temperature $\tau'$ is applied to normalize $\dot{S} \in [-1, 1]$ into $\hat{S} \in [0, 1]$:
\begin{equation}
	\hat{S} = \delta(\dot{S}) = \{\frac{\exp(\dot{S}^c_i/\tau')}{\sum_{d\in C} \exp(\dot{S}^d_i / \tau')}\ |\ c\in C, i=1,2,...,\mathrm{HW}\}.
	\label{eq:Normalize}
\end{equation}

\begin{figure}
	\centering
	\includegraphics[width=1.0\linewidth]{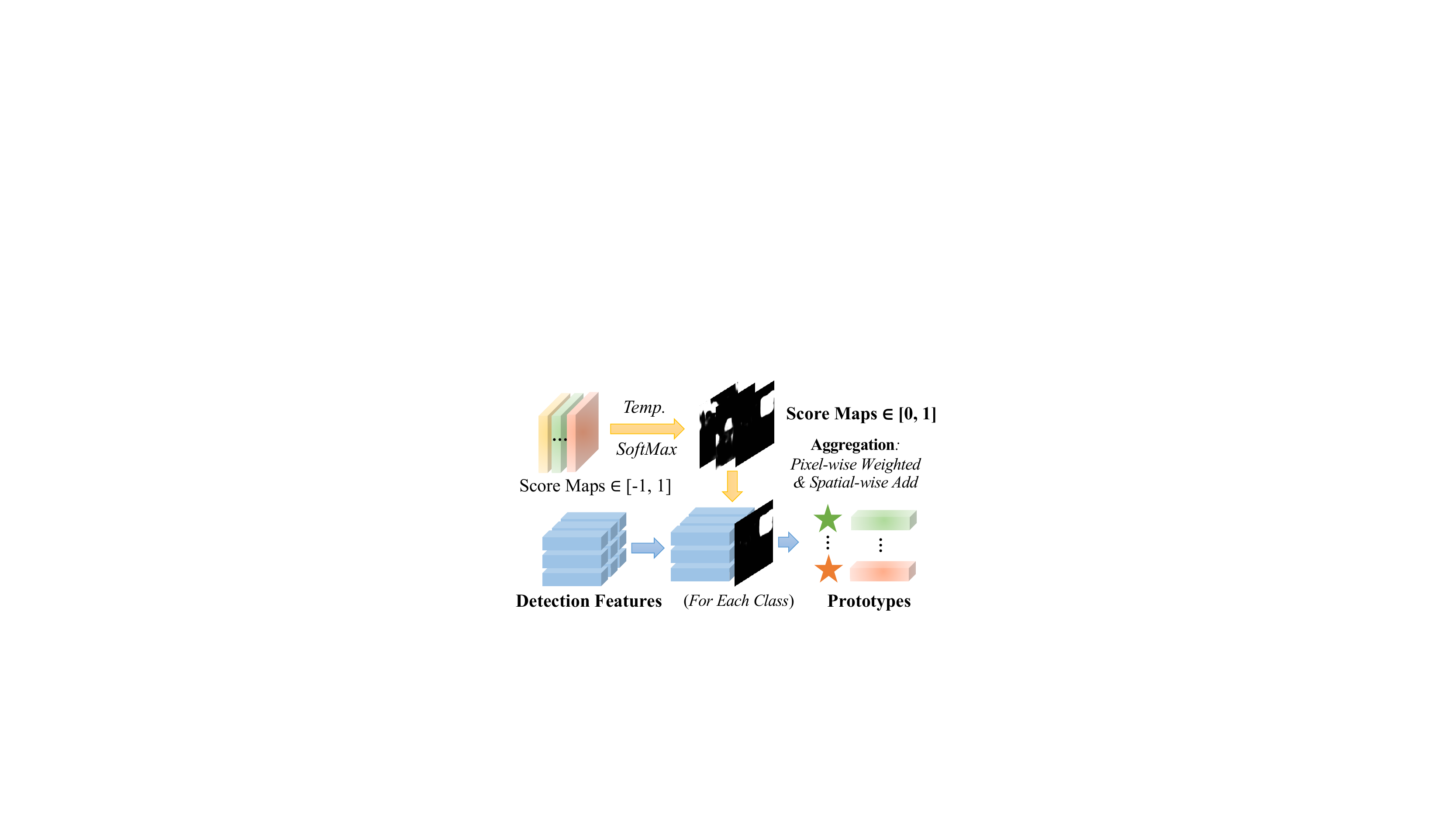}
	
	\caption{Pixel-wise Aggregation for our proposed PSC. ``Detection Features'' are pixel-wise weighted by predicted score maps of VLS and then spatially added into prototypes as an aggregation.}
	\label{fig:Aggregation}
\end{figure}

Since we should not disturb the self-supervised learning of VLS and only aim to improve the detection, ``Detection Features'' $E \in \mathbb{R}^{\rm D\times H \times W}$ are supervised. As shown in Figure \ref{fig:Aggregation}, prototypes of each class are obtained via aggregating $E$ pixel-wisely by $\hat{S}$, denoted as Pixel-wise Aggregation:
\begin{equation}
	P^{-} =  E \cdot \hat{S} = \{P^{c-}=\sum\nolimits_{i} E_i \cdot \hat{S}^c_i\ |\ c \in C\},
	\label{eq:Prototype}
\end{equation} 
where $\cdot$ is matrix multiplication, and $P^{-} \in \mathbb{R}^{\rm D\times (N-1)}$ are  negative prototypes of $\mathrm{N}-1$ non-human classes with channels $\mathrm{D}$. Weighted by $\hat{S}^c_i \in \mathbb{R}$, feature $E_i\in \mathbb{R}^{D}$ at each position $i=1,2,...,\mathrm{HW}$ is aggregated. Similarly, 2D Gaussians of pedestrian positions $G \in \mathbb{R}^{\rm H\times W \times 1}$ from the baseline CSP \cite{liu2019high} can replace the $\hat{S}$ for positive prototype $P^+$.

Finally, each pixel-wise $E_j$ of the annotated pedestrians is supervised as ``query'' of contrastive loss function $\mathcal{L}_{PSC}$ in Figure \ref{fig:Teaser}(d) and \ref{fig:Pipeline}(b), which pulls them close to $P^+$ and pushes them away from each $P^{c-}$. $E_j$ is located by the $>0$ positions $j$ of $G$ and $|G^{>0}|=\mathrm{M}$. $\mathcal{L}_{PSC}$ is formulated as: 
\begin{equation}
	- \frac{1}{\mathrm{M}} \sum_{j\in G^{>0}} \log \frac{\exp( E_j \cdot P^+ /\tau)}{\exp( E_j \cdot P^+ /\tau)+\sum_{c,b} \exp( E_j \cdot P^{c-}_{b} /\tau)},
	\label{eq:PSC}
\end{equation}
where self-normalization of the features and prototypes, e.g., $E_j / \Vert E_j \Vert$, is omitted for simplicity. Similar to some contrastive pretraining methods \cite{he2020momentum,chen2020improved}, negative prototypes are expanded to all images $b\in B$ inside the mini-batch $B$. 

In brief, based on the contrastive self-supervision, PSC trains the detector to better discriminate the pedestrians and other classes via the positive and negative prototypes, owing to the explicit and semantic contexts obtained from VLS. By the integration of the VLS and PSC, our proposed VLPD is supervised by $\mathcal{L}_{Det}$ \cite{liu2019high}, $\mathcal{L}_{VLS}$ and $\mathcal{L}_{PSC}$ simultaneously: 
\begin{equation}
	\mathcal{L} = \mathcal{L}_{Det} + \lambda_1\mathcal{L}_{VLS} + \lambda_2\mathcal{L}_{PSC}.
\end{equation}

\subsection{Detection Head}

Following the anchor-free style of our baseline CSP \cite{liu2019high}, Detection Head in Figure \ref{fig:Pipeline} and \ref{fig:Details} firstly decreases the channels of ``Detection Features'' via convolution layers. Then multiple branches predict result maps: ``Center Heatmap'' to classify the centers of pedestrians, ``Scale Map'' to predict heights with a fixed aspect ratio 0.41 for widths, and ``Offset Map'' to adjust the localization horizontally and vertically. Finally, these maps are assembled into bounding boxes of pedestrians. More details can be found in the CSP paper.

\section{Experiments}

In this section, extensive experiments are conducted on two popular benchmarks for pedestrian detection, i.e., Caltech and CityPersons, to evaluate our proposed VLPD method. Ablation study is performed on key components VLS and PSC. Furthermore, we also report the state-of-the-art comparisons on both benchmarks. For more experiments and visualizations, please refer to supplementary materials.

\subsection{Datasets}

The Caltech pedestrian dataset \cite{dollar2009pedestrian} comprises 2.5-hour video data captured on the urban areas of Los Angeles, with 4024 images for testing.  Over 70\% of pedestrians are less than 100 pixels high, including particularly small pedestrians that are less than 50 pixels. By fixing the inconsistency and box misalignment, Zhang et. al. \cite{zhang2016far} has released a new version of the annotations. For fair comparisons, all the following evaluations are performed based on the new version.

CityPersons \cite{zhang2017citypersons} is a more recently published large-scale pedestrian detection dataset. 2975 images are split for training and 500 images for validation. The standard evaluation metric is as follows: log miss rate is averaged over the false positive per image (FPPI) $\in [10^{-2};100]$, denoted as $\mathrm{MR}^{-2}$. All tests are applied on the original data (1$\times$) without resizing and any extra visible or head labels for fair comparisons.

\begin{figure*}
	\centering
	\includegraphics[width=0.99\textwidth]{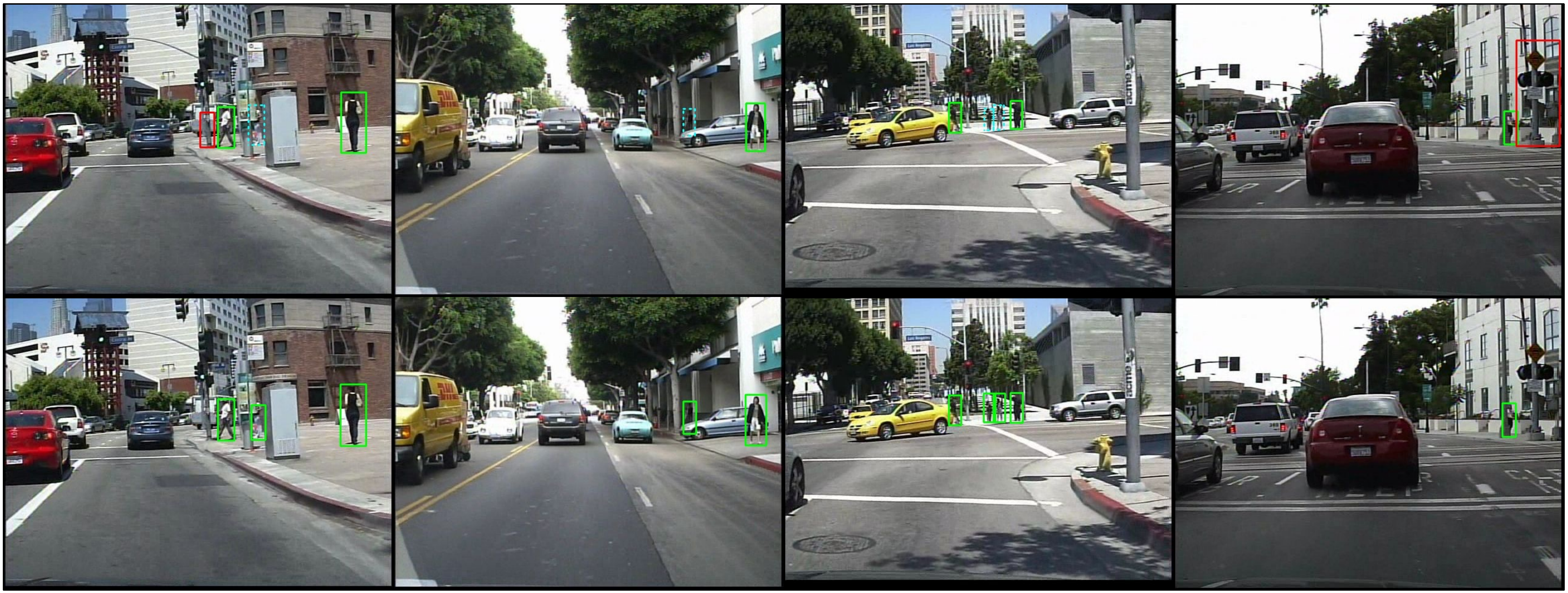}
	
	\caption{Qualitative analysis on Caltech \cite{dollar2009pedestrian} between the baseline CSP \cite{liu2019high} (top) and our proposed VLPD (bottom). Green are correct detections, red are wrong detections, and dashed blue are missing detections. With the powerful vision-language semantic self-supervision, our proposed VLPD is context-aware and thus more robust to human-like objects,  inter-class occlusion and ambiguous small pedestrians.}
	\label{fig:BboxVisualize}
\end{figure*}

\subsection{Implementation Details}

Our proposed method is based on a powerful pedestrian detector CSP \cite{liu2019high}, which is re-implemented on PyTorch \cite{paszke2019pytorch} framework from the original Keras one. Adam \cite{kingma2014adam} is adopted for optimization. The backbone network is ResNet-50 \cite{he2016deep} pretrained on ImageNet \cite{deng2009imagenet} by fully-supervised image classification or WIT \cite{radford2021learning} by self-supervised vision-language contrastive learning. For Caltech, one Nvidia 3090 GPU is utilized for training with $10^{-4}$ learning rate. For CityPersons, two 3090 GPUs are used with $2\times 10^{-4}$. Batch sizes are set following \cite{liu2019high}. All tests are conducted on a single 3090 GPU. The size of training images is 336$\times$448 for Caltech and 640$\times$1280 for CityPersons. For our proposed VLS, its loss weight $\lambda_1=100$. For PSC, its weight $\lambda_2=10^{-4}$ for Caltech and $10^{-3}$ for CityPersons. Temperatures $\tau'=10^{-3}$ and $\tau=7\times10^{-2}$ following \cite{he2020momentum}.

\begin{table}
	\centering
	\caption{Overall ablation study for key components of our proposed VLPD, including VLS and PSC. \textbf{Bolden} are the best results.}
	\begin{tabular}{c|c|c|c}
		\toprule [1.2pt]
		Method & Reasonable & Small & HO  \\ 
		\midrule [0.8pt]
		CSP \cite{liu2019high} & 11.0 & 16.0 & - \\
		CSP (our re-imp.) & 10.96 & 16.05 & 40.59 \\
		\midrule [0.8pt]
		CSP w/ CLIP & 10.13 & 12.59 & 38.97 \\
		+VLS & 9.70 & 12.57 & 36.50 \\
		+VLS+PSC\textbf{=VLPD} & \textbf{9.41} & \textbf{10.93} & \textbf{34.88} \\
		\bottomrule [1.2pt]
	\end{tabular}
	\label{table:OverallAblation}
\end{table}

\subsection{Ablation Study}

The ablation study is firstly performed on the popular CityPersons dataset. Comprehensive subset Reasonable, more challenging ones Small and HO (Heavy Occlusion, visible rate $\in [0.2, 0.65]$) are widely-used for comparisons. 

Table \ref{table:OverallAblation} illustrates the overall ablation study for each key components of our proposed VLPD. We provide the original results of CSP\cite{liu2019high} and our re-implemented one by PyTorch\cite{paszke2019pytorch}. Under the CLIP \cite{radford2021learning} initialized visual encoder as a precondition of VLS, the improvements are limited because merely vision-language pretraining cannot fully handle the context modeling. With VLS as well as PSC based on VLS, our proposed VLPD gains significant boosts especially on the context-related subsets Small and HO. 

In Table \ref{table:VLSAblation}, different policies of the class set for our proposed VLS are evaluated. Full CityScapes policy adopts all the classes of the CityScapes dataset \cite{cordts2016cityscapes}, and Full Compacted uses $2^\mathrm{nd}$ level of classes. Both the too scattered and concentrated sets lead to performance losses. Instead, our proposed policy for VLS handles the largest ``vehicle'' classes via frequency statistics for better context modeling. 

Meanwhile, sub-items of this policy are evaluated at the bottom of Table \ref{table:VLSAblation} via recovering to the $1^{\rm st}$ column of Table \ref{table:Policy}. Table \ref{table:PSCAblation} shows that our PSC in Eq.\ref{eq:PSC} with cross-image negatives and inner-image positives of ``Detection Features'' $E$ without disturbing the $\dot{S}$ from VLS performs the best. 

\begin{table}
	\centering
	\caption{Different policies of the class set for our proposed VLS.}
	\begin{tabular}{c|c|c|c}
		\toprule [1.2pt]
		Method & Reasonable & Small & HO  \\ 
		\midrule [0.8pt]		
		CSP w/ CLIP & 10.13 & 12.59 & 38.97 \\
		\midrule [0.8pt]
		+ Full CityScapes & 10.40 & 12.87 & 37.14 \\
		+ Full Compacted & 10.47 & 13.30 & 40.49 \\
		\textbf{+ VLS (ours)} & \textbf{9.70} & 12.57 & \textbf{36.50} \\
		\midrule [0.8pt]
		w/o ground  & 10.51 & 13.53 & 37.48 \\
		w/o building  & 10.42 & 12.84 & 37.70 \\
		w/o tree  & 10.34 & 12.66 & 38.39 \\
		w/o human & 10.24 & 12.52 & 38.47 \\
		\makecell[c]{w/o \{car, bicycle,\\ bus, truck\}} & 10.61 & 13.61 & 38.84 \\
		w/o traffic sign & 10.11& \textbf{12.27} & 37.22 \\
		\bottomrule [1.2pt]
	\end{tabular}
	\label{table:VLSAblation}
\end{table}

\begin{table}
	\centering
	\caption{Different prototypes and features for our proposed PSC.}
	\begin{tabular}{c|c|c|c}
		\toprule [1.2pt]
		Method & Reasonable & Small & HO  \\ 
		\midrule [0.8pt]
		\textbf{VLPD (w/ PSC, ours)}  & \textbf{9.41} & \textbf{10.93} & \textbf{34.88} \\
		\midrule [0.8pt]
		Cross $\rightarrow$ Inner-img Neg. & 10.12 & 12.67 & 37.90 \\
		Inner $\rightarrow$ Cross-img Pos. & 10.58 & 13.14 & 38.08 \\
		\midrule [0.8pt]
		$E$ $\rightarrow$ $\mathrm{Concate}(E, \dot{S})$ & 10.04& 12.93 & 38.21 \\
		\bottomrule [1.2pt]
	\end{tabular}
	\label{table:PSCAblation}
\end{table}

\subsection{Comparisons with the State-of-the-arts}

For CityPersons, we compare our proposed VLPD with various state-of-the-art methods: AMSCNN\cite{zhang2020asymmetric}, DHRNet\cite{ding2021learning} and SML\cite{wu2020self} for scale-variation; RepLoss\cite{wang2018repulsion}, Adaptive NMS\cite{liu2019adaptive}, PBM+R$^2$NMS\cite{huang2020nms}, CaSe\cite{xie2020count}, NMS-Ped\cite{luo2021nms} and MAPD\cite{wang2021mapd} for intra-class occlusion; OR-CNN\cite{zhang2018occlusion}, HBAN\cite{lu2020semantic} and PRNet++\cite{song2022prnet++} for part-aware occlusion handling; LBST\cite{cao2019taking}, ALFNet\cite{liu2018learning}, CSP\cite{liu2019high}, AP$^2$M\cite{liu2021adaptive} and BGCNet\cite{li2020box} for generic hard pedestrian detection. Note that the context-related methods are: SMPD\cite{jiang2022urban} with segmentation annotation, EGCL\cite{lin2022pedestrian} with proposal-wise contrastive learning and FC-Net\cite{zhang2020feature} with neighbor region modeling. As illustrated in Table \ref{table:SOTACityPersons}, our VLPD out-performs them comprehensively among all the subsets. 

\begin{table}
	\centering
	\caption{Comparison with the state-of-the-arts on CityPersons.}
	\begin{tabular}{c|c|c|c|c|c}
		\toprule [1.2pt]
		Methods & R & Hea. & Partial & Bare & Small \\
		\midrule [0.8pt]
		AMSCNN\cite{zhang2020asymmetric} & 14.0 & - & - & - & 12.6 \\
		FC-Net\cite{zhang2020feature} & 13.9 & 46.8 & - & - & - \\
		RepLoss\cite{wang2018repulsion} & 13.2 & 56.9 & 16.8 & 7.6 & 42.6 \\
		OR-CNN\cite{zhang2018occlusion} & 12.8 & 55.7 & 15.3 & 6.7 & 42.3 \\
		LBST\cite{cao2019taking} & 12.6 & 48.7 & 18.6 & - & - \\ 
		SML\cite{wu2020self} & 12.3 & - & - & - & 19.3 \\ 
		ALFNet\cite{liu2018learning} & 12.0 & 51.9 & 11.4 & 8.4 & 19.0 \\  
		AdaNMS\cite{liu2019adaptive} & 11.9 & 55.2 & 12.6 & 6.2 & - \\
		PR$^2$NMS\cite{huang2020nms} & 11.1 & 53.3 & - & - & - \\ 
		CSP\cite{liu2019high} & 11.0 & 49.3 & 10.4 & 7.3 & 16.0 \\ 
		CaSe\cite{xie2020count} & 11.0 & 50.3 & - & - & - \\ 
		HBAN\cite{lu2020semantic}  & 10.9 & 47.0 & - & - & - \\  
		EGCL\cite{lin2022pedestrian} & 10.9 & 46.4 & 11.6 & 7.4 & - \\ 
		PRNet++\cite{song2022prnet++} & 10.7 & 51.2 & 9.9 & 6.9 & - \\ 
		AP$^2$M\cite{liu2021adaptive} & 10.4 & 48.6 & 9.7 & 6.2 & 15.3 \\ 
		DHRNet\cite{ding2021learning} & 10.4 & - & - & - & 13.4 \\ 
		NMS-Ped\cite{luo2021nms} & 10.1 & - & - & - & - \\ 
		SMPD\cite{jiang2022urban} & 9.9 & 45.6 & 9.0& 6.5 & - \\ 
		MAPD\cite{wang2021mapd} & 9.7 & 46.4 & 9.9 & 6.1 & - \\ 
		BGCNet\cite{li2020box} & 9.4 & 45.9 & 9.0 & 6.4 & - \\ 
		\midrule [0.8pt]
		\textbf{VLPD (ours)} & \textbf{9.4} & \textbf{43.1} & \textbf{8.8} & \textbf{6.1} & \textbf{10.9} \\ 
		\bottomrule [1.2pt]
	\end{tabular}
	\label{table:SOTACityPersons}
\end{table}

In details, we denote Reasonable subset as ``R'' and the occlusion one Heavy (visible rate $\in [0, 0.65]$) as ``Hea.'' in Table \ref{table:SOTACityPersons}. We also compare our proposed VLPD with the methods on other occlusion subsets in Table \ref{table:OccCityPersons}, i.e., R+HO (Reasonable+HO, visible rate $\in [0.2, 1]$) and HO. Our method keeps the best under these setting changes. 

For Caltech, additional state-of-the-art methods are compared: AR-Ped \cite{brazil2019pedestrian} for generic hard pedestrian handling; JointDet \cite{chi2020relational}, PedHunter \cite{chi2020pedhunter} and DMSFLN \cite{he2021occluded} with visible or head labels. In Table \ref{table:SOTACaltech}, without any extra labels, our proposed method VLPD also surpasses them significantly. Its Reasonable 2.27\% is better than 2.31\% of \cite{chi2020pedhunter}. Context-related challenging subsets Heavy Occlusion 37.7\% and All 52.4\% are especially better than other methods.

In conclusion, our proposed VLPD has become a new state-of-the-art on both benchmarks especially in context-related subsets, which sufficiently validates its power of vision-language semantic self-supervision to explicitly model semantic contexts without any extra labels and better discriminate pedestrians from other contextual classes.

\begin{table}
	\centering
	\caption{Comparison on other occlusion subsets on CityPersons.}
	\begin{tabular}{c|c|c|c}
		\toprule [1.2pt]
		Methods & Reasonable & R+HO & HO \\
		\midrule [0.8pt]
		FC-Net\cite{zhang2020feature} & 13.9 & 29.6 & - \\ 
		ALFNet\cite{liu2018learning} & 12.0 & 26.3 & 43.8 \\ 
		EGCL\cite{lin2022pedestrian} & 10.9 & 24.8 & 39.3 \\ 
		PRNet++\cite{song2022prnet++} & 10.7 & 25.4 & 40.9 \\ 
		SMPD\cite{jiang2022urban}  & 9.9 & - & 36.6 \\ 
		\midrule [0.8pt]
		\textbf{VLPD (ours)} & \textbf{9.4} & \textbf{21.7} & \textbf{34.9} \\ 
		\bottomrule [1.2pt]
	\end{tabular}
	\label{table:OccCityPersons}
\end{table}

\begin{table}
	\centering
	\caption{Comparison with the state-of-the-arts on Caltech.}
	\begin{tabular}{c|c|c|c}
		\toprule [1.2pt]
		Methods & Reasonable & All & Heavy \\
		\midrule [0.8pt]
		ALFNet\cite{liu2018learning} & 6.1 & 59.1 & 51.0 \\ 
		RepLoss\cite{wang2018repulsion} & 5.0 & 59.0 & 47.9 \\ 
		CSP\cite{liu2019high} & 4.5 & 56.9 & 45.8 \\ 
		AR-Ped\cite{brazil2019pedestrian} & 4.4 & - & - \\ 
		BGCNet\cite{li2020box} & 4.1 & - & 42.0 \\ 
		DHRNet\cite{ding2021learning} & 3.4 & - & - \\ 
		AP$^2$M\cite{liu2021adaptive} & 3.3 & 55.9 & 42.2 \\ 
		JointDet\cite{chi2020relational} & 3.0 & - & - \\ 
		DMSFLN\cite{he2021occluded} & 2.7 & - & - \\ 
		PedHunter\cite{chi2020pedhunter} & 2.3 & - & - \\
		\midrule [0.8pt]
		\textbf{VLPD (ours)} & \textbf{2.3} & \textbf{52.4 }& \textbf{37.7} \\ 
		\bottomrule [1.2pt]
	\end{tabular}
	\label{table:SOTACaltech}
\end{table}

\section{Conclusion}

In this paper, we have proposed a novel pedestrian detection method VLPD for explicit contexts modeling towards challenging problems, e.g., human-like objects and small scale or heavily occluded pedestrians. It tackles these challenges via vision-language semantic self-supervision with two key components: VLS is proposed to leverage vision-language models to recognize semantic classes for explicit contexts, which learns fully-supervised pedestrian detection and self-supervised segmentation via pseudo labels by cross-modal mapping. PSC is proposed to adopt contrastive self-supervision for better discriminating pedestrians and semantic classes based on explicit contexts from VLS. By the integration of VLS and PSC, our VLPD achieves the new cutting-edge performances on two challenging benchmarks Caltech and CityPersons, especially on the very difficult circumstances of small scale and heavy occlusion. 

\section{Acknowledgments}

This work was supported by National Key Research and Development Program of China (2020AAA0109701), National Natural Science Foundation of China (62072032, 62076024), and National Science Fund for Distinguished Young Scholars (62125601). 

{\small
	\bibliographystyle{ieee_fullname}
	\bibliography{egbib}
}

\end{document}